\documentclass[journal]{IEEEtran}
\usepackage{times}
\usepackage{ifpdf}
\usepackage[pdftex]{graphicx}
\usepackage{amsmath}
\usepackage{amssymb}
\usepackage{subfigure}
\usepackage[ruled,commentsnumbered,noline]{algorithm2e}
\usepackage{paralist}
\usepackage{multirow}
\usepackage{slashbox}
\usepackage{cite}
\usepackage{algorithmic}
\usepackage{bm}
\usepackage{amsfonts}
\usepackage{graphicx}
\usepackage{textcomp}
\usepackage{CJK}
\usepackage{indentfirst}
\usepackage{makecell}
\usepackage{array}
\usepackage{float}
\usepackage{stfloats}
\usepackage{makecell}
\usepackage{amsthm,amsmath,amssymb}
\usepackage{mathrsfs}
\usepackage{color}
\usepackage{listings}
\usepackage{xcolor} %
\usepackage{pifont}  %

\usepackage{booktabs}  %
\usepackage{hhline}

\begin{document}

\title{AttDiff-GAN: A Hybrid Diffusion-GAN Framework for Facial Attribute Editing}


\author{Wenmin~Huang, Weiqi~Luo, ~\IEEEmembership{Senior Member,~IEEE,} Xiaochun~Cao, ~\IEEEmembership{Senior Member,~IEEE,} Jiwu~Huang, ~\IEEEmembership{Fellow,~IEEE}

\thanks{This work was supported  in part by the National Natural Science Foundation of China under Grants  62472458. (Corresponding author: Weiqi~Luo.)}

\thanks{Wenmin Huang and Weiqi Luo are with GuangDong Province Key Lab of Information Security Technology and School of Computer Science and Engineering, Sun Yat-sen University, Guangdong 510000, China (E-mail: huangwm36@mail2.sysu.edu.cn; luoweiqi@mail.sysu.edu.cn).

Xiaochun Cao is with School of Cyber Science and Technology, Shenzhen Campus, Sun Yat-sen University, Shenzhen 518107, China (E-mail: caoxiaochun@mail.sysu.edu.cn).

Jiwu Huang is with the Guangdong Laboratory of Machine Perception and  Intelligent Computing, Faculty of Engineering, Shenzhen MSU-BIT University, Shenzhen, 518116, China (E-mail: jwhuang@smbu.edu.cn).}

}

\markboth{}%
{Shell \MakeLowercase{\textit{et al.}}: Bare Demo of IEEEtran.cls for IEEE Journals}

\maketitle
\begin{abstract}
Facial attribute editing aims to modify target attributes while preserving attribute-irrelevant content and overall image fidelity. Existing GAN-based methods provide favorable controllability, but often suffer from weak alignment between style codes and attribute semantics. Diffusion-based methods can synthesize highly realistic images; however, their editing precision is limited by the entanglement of semantic directions among different attributes. 
In this paper, we propose AttDiff-GAN, a hybrid framework that combines GAN-based attribute manipulation with diffusion-based image generation. A key challenge in such integration lies in the inconsistency between one-step adversarial learning and multi-step diffusion denoising, which makes effective optimization difficult. 
To address this issue, we decouple attribute editing from image synthesis by introducing a feature-level adversarial learning scheme to learn explicit attribute manipulation, and then using the manipulated features to guide the diffusion process for image generation, while also removing the reliance on semantic direction-based editing.  Moreover, we enhance style–attribute alignment by introducing PriorMapper, which incorporates facial priors into style generation, and RefineExtractor, which captures global semantic relationships through a Transformer for more precise style extraction. Experimental results on CelebA-HQ show that the proposed method achieves more accurate facial attribute editing and better preservation of non-target attributes than  state-of-the-art methods in both qualitative and quantitative evaluations.
\end{abstract}

\IEEEpeerreviewmaketitle
\section{Introduction}\label{sec1}
\begin{figure}[!t]
\centering
\includegraphics[width=3.4in]{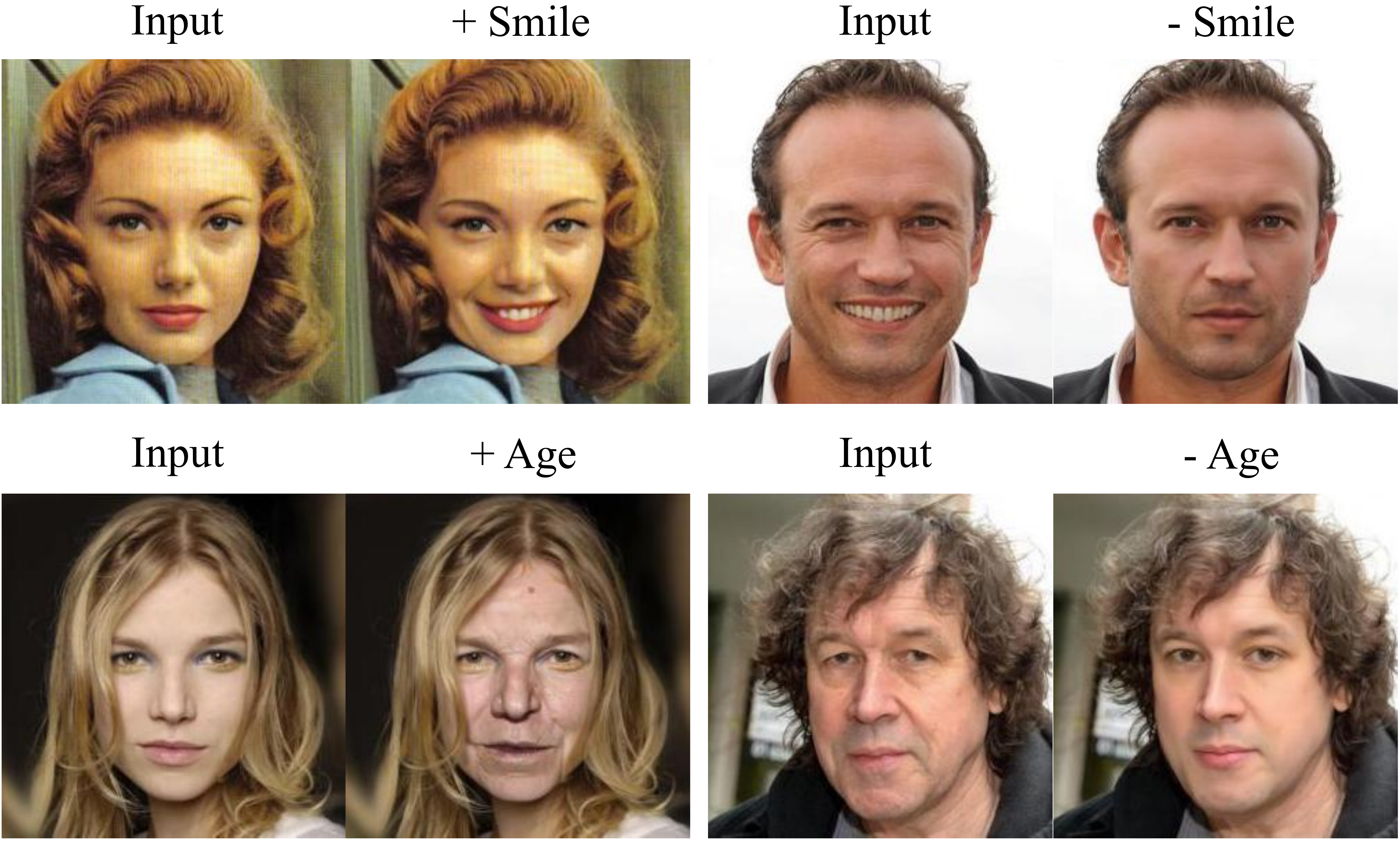}
\caption{Illustration of facial attribute editing with the proposed AttDiff-GAN.}
\label{fig_1}
\end{figure}

\IEEEPARstart{F}{acial} attribute editing aims to perform controlled modifications on specific attributes of a face image, such as adding or removing a smile, altering perceived age, and more, as shown in Fig.~\ref{fig_1}. It plays a crucial role in applications such as virtual avatar creation, professional photo editing, and data augmentation for facial recognition tasks, where precise and reliable manipulation is essential. Ideally, only the target attribute is modified, without introducing unintended changes. However, achieving such isolated editing remains a significant challenge due to the strong correlations among facial attributes, where alterations in one attribute often cause unintended modifications or visual inconsistencies.

To address this challenge, numerous methods based on advanced conditional generative adversarial networks (cGANs) have been proposed. These methods can be broadly categorized into two groups: those based on pre-trained GANs and those that directly learn image-to-image translation.
The first group~\cite{WangCVPR2022,PehlivanCVPR2023} focuses on manipulating the latent space of pre-trained GANs. Specifically, the input image is first projected into the latent space via inversion, and attribute editing is then achieved by moving its latent code along predefined semantic directions.
The second group formulates attribute editing as an image-to-image translation problem. In this context, some methods~\cite{HeTIP2019,HuangTCSVT2024} learn a direct mapping between simple binary attribute vectors and facial appearances, enabling attribute editing by modifying the attribute vectors. 
More advanced approaches~\cite{DalvaTPAMI2023,HuangAAAI2024} represent attributes as expressive style codes, which can be flexibly obtained either from random noise via a mapper or from reference images via an extractor. By manipulating images based on these style codes, flexible and controllable attribute editing can be achieved, making this approach a focal point of recent research.
However, precise attribute editing requires effective alignment between style codes and attribute semantics. Existing methods~\cite{DalvaTPAMI2023, HuangAAAI2024, ren2025facial} rely on MLP-based mappers and CNN-based extractors, which either lack facial priors or fail to capture global context, resulting in inadequate understanding of facial semantics and poor alignment between style codes and attribute semantics.

Recently, diffusion models (DMs) have achieved great success in generating high-quality and diverse images, inspiring numerous works on facial tasks such as identity-driven generation\cite{RuizCVPR2024}, stroke-based editing\cite{MengICLR2021}, and face restoration\cite{wang2025osdface}. In facial attribute editing, several works\cite{PreechakulCVPR2022, KimCVPR2023}, inspired by pre-trained GAN-based methods, map images into the latent space of a conditional diffusion model and perform editing by manipulating the latent codes along semantic directions, which are then used to guide the diffusion process for image generation.
Although these approaches leverage the powerful generation ability of diffusion models, the coupling of semantic directions across attributes makes them difficult to preserve irrelevant attributes and details during editing.

In this paper, we study how to combine the controllability of GANs with the high-quality image generation of diffusion models for facial attribute editing. A key challenge lies in the inconsistency between one-step adversarial learning in GANs and the multi-step denoising process in diffusion models. Directly propagating adversarial supervision through the entire denoising trajectory incurs prohibitive computational cost and leads to unstable optimization. 
To address this, we propose AttDiff-GAN, a hybrid framework that decouples attribute editing from image generation. The core idea behind AttDiff-GAN is to perform attribute editing in a feature space using GANs and then use the edited features to guide the diffusion model for image synthesis. 
To facilitate this integration, we design a feature-level adversarial learning scheme where a discriminator is used to distinguish real features from edited ones, allowing the generator to explicitly learn attribute manipulation. This design fundamentally removes the reliance on semantic direction-based editing.
Additionally, we improve style–attribute alignment by introducing PriorMapper, which incorporates facial priors from the input image into the style code generation process, thus avoiding reliance on purely random noise, and RefineExtractor, which leverages Transformer-based global context modeling for more accurate style extraction. In summary, the main contributions of this paper are as follows:

\begin{itemize}
\item We propose AttDiff-GAN, which resolves the optimization inconsistency between GANs and diffusion models by decoupling attribute editing from image generation and introducing feature-level adversarial training to achieve controllable, high-fidelity editing results.  
\item We improve style-attribute alignment with PriorMapper, which incorporates facial priors into style code generation, and RefineExtractor, which leverages Transformer-based context modeling for accurate style extraction.
\item Extensive experiments on CelebA-HQ~\cite{KarrasICLR2018} demonstrate that our method achieves precise control over target attributes, and outperforms state-of-the-art methods in both qualitative and quantitative evaluations.
\end{itemize}

\section{Related Work}\label{sec2}

\subsection{Deep Generative Models}\label{sec2.1}
Deep generative models, including Variational Autoencoders (VAEs) \cite{KingmaarX2013}, normalizing flows \cite{RezendearX2015}, and autoregressive models\cite{van2016pixel}, have revolutionized image synthesis, with GANs and diffusion models being the most prominent approaches.
GANs\cite{GoodfellowNIPS2014} consist of a generator and a discriminator. The generator produces realistic images, while the discriminator distinguishes between real and generated images. Through this adversarial process, the generator learns to create high-quality images. GANs have been widely applied in tasks such as facial image generation\cite{KarrasCVPR2020}, image super-resolution \cite{10902474}, and style transfer \cite{10198457}, thanks to their controllability and ability to generate high-resolution images.
Diffusion models\cite{HoNIPS2020,SongICLR2021} generate images by progressively adding noise to data and learning to reverse this process. Diffusion models have shown strong performance in generating high-quality images and are increasingly applied in  face restoration\cite{wang2025osdface}, text-to-image generation\cite{rombach2022high}, and instruction-based image editing\cite{brooks2023instructpix2pix,11363596}.
In this paper, we propose a hybrid Diffusion-GAN framework that combines the control of GANs with the stable, high-quality generation of diffusion models.

\subsection{Facial Attribute Editing}\label{sec2.2}
\noindent \textbf{GAN-based Facial Attribute Editing.}
Facial attribute editing has long been dominated by GAN-based methods, which can be broadly categorized into two types: manipulating the latent space of pre-trained GANs \cite{WangCVPR2022, PehlivanCVPR2023} and learning image-to-image translation \cite{GaoCVPR2021, HuangTCSVT2024, DalvaTPAMI2023, HuangAAAI2024}.
The first method leverages the structured latent space of GAN models, especially StyleGAN \cite{KarrasCVPR2020}, where each semantic attribute corresponds to a specific direction in the latent space\cite{KarrasCVPR2019, ShenTPAMI2020}. By performing GAN inversion, i.e., mapping facial images into the GAN's latent space, and modifying the latent codes along predefined directions, facial attributes can be edited effectively. While this approach offers convenience, such as avoiding the need to retrain the GAN, challenges arise in aligning latent codes with the true distribution of the GAN's latent space, often resulting in loss of detail or lower editing accuracy\cite{WangCVPR2022,PehlivanCVPR2023}. Additionally, semantic directions are typically not orthogonal, and this inherent coupling can cause unintended modifications during editing\cite{PatashnikCVPR2021,WuCVPR2021}.

The second method treats facial attribute editing as an image-to-image translation task, typically without paired images. 
Early works \cite{LiarX2016, ShenCVPR2017} trained encoder-decoder networks on database divided by specific attributes. 
Later methods \cite{GaoCVPR2021, HuangTCSVT2024} introduce conditional GANs that use binary attribute vectors, with each element corresponding to a specific attribute, as control conditions to guide the editing process. This enables attribute editing by modifying the attribute vector, allowing a single model to support multiple attribute edits.
Additionally, recent GAN-based approaches\cite{DalvaTPAMI2023, HuangAAAI2024, ren2025facial} incorporate style codes for more controllable attribute editing. Specifically, style codes are either generated from random noise via a mapper (latent-guided) or extracted from reference images via an extractor (reference-guided). The generator then retrieves the corresponding style code based on the target label to perform the attribute editing. In this paper, we build upon this editing strategy and enhance style code generation and extraction by  incorporating
facial priors into the mapper and Transformers\cite{dosovitskiy2020image} into the extractor, achieving precise editing results.

\noindent \textbf{Diffusion-based Facial Attribute Editing.}
Recently, diffusion models \cite{HoNIPS2020, SongICLR2021} have demonstrated considerable potential in facial image manipulation tasks.
In particular, the stable training and high-quality generation abilities of diffusion models provide an attractive alternative for facial attribute editing. A notable approach is DiffAE \cite{PreechakulCVPR2022}, which integrates a  encoder within the DDIM framework \cite{SongICLR2021} to learn a semantically rich latent space, similar to GANs. Attribute editing is achieved by adjusting latent codes along semantic directions defined by a linear classifier trained on images with and without the target attribute.
Building on DiffAE, DiffVAE \cite{KimCVPR2023} extends this framework by incorporating an identity encoder and landmark encoder, enabling video-based facial attribute editing while ensuring temporal consistency across frames.
Although these methods leverage the strengths of diffusion models for high-quality image generation, their editing process still relies on predefined semantic directions. The fixed and coupled nature of semantic directions limits the model’s ability to edit individual attributes without affecting other attributes or content. To address this limitation, we decouple attribute editing from image generation and introduce feature-level adversarial training to directly model attribute manipulation, enabling controllable, high-quality editing.

\begin{figure*}[!t]
\centering
\includegraphics[width=7.1in]{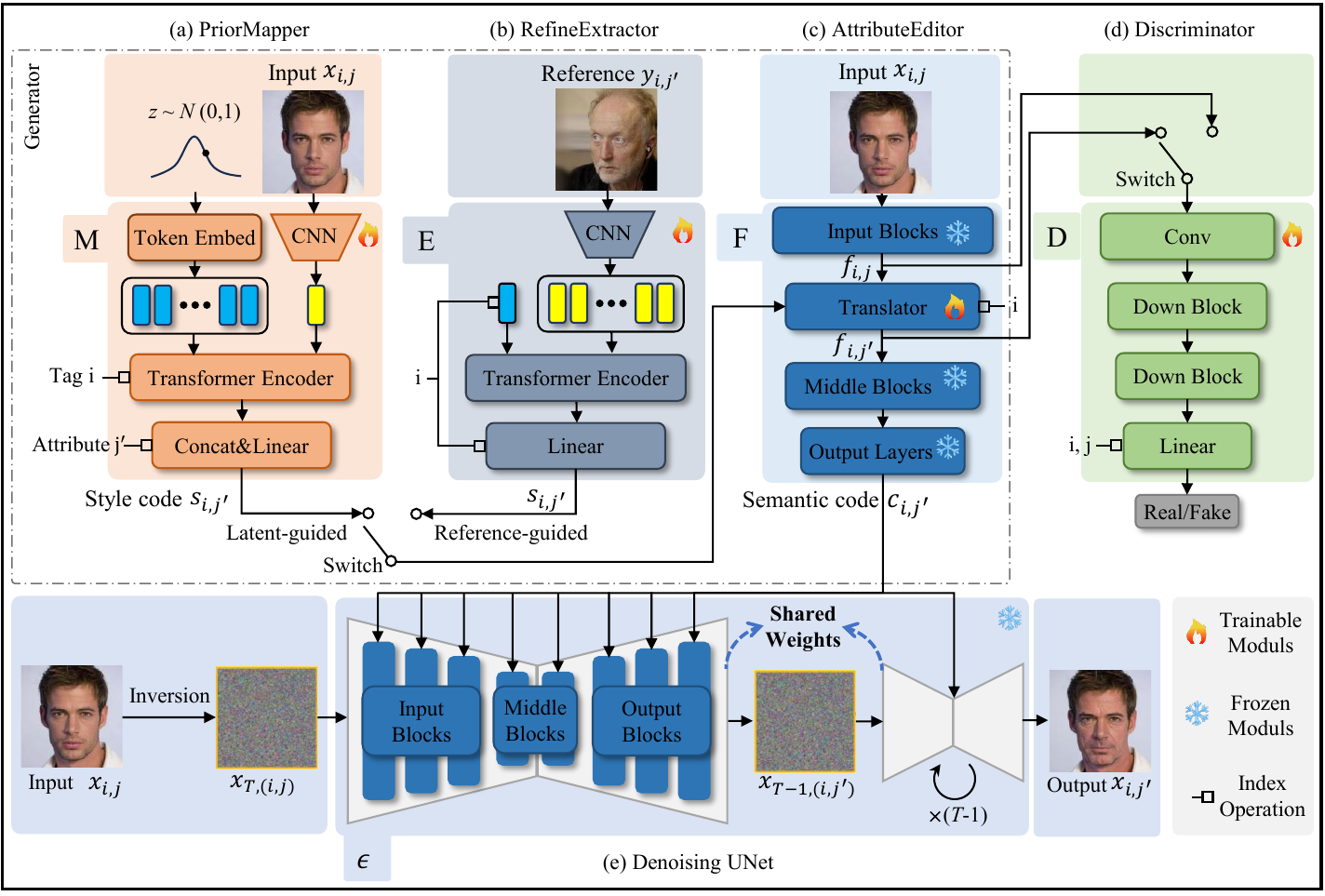}
\caption{The overall framework of AttDiff-GAN consists of a feature-level generator and discriminator $D$ jointly designed for attribute editing, along with a diffusion-based denoising U-Net $\epsilon$ for image generation, where the generator is composed of PriorMapper $M$, RefineExtractor $E$, and AttributeEditor $F$. }
\label{fig_2}
\end{figure*}


\section{Proposed Method}\label{sec3}
For clarity, we adopt the hierarchical labeling scheme from \cite{LiCVPR2021} to represent the relationship between attributes. Tag \( i \) denotes an abstract attribute category, while \( j \) represents the specific attribute under tag $i$. For example, when \( i \) represents ``age'', \( j \) can correspond to specific attributes like ``young'' or ``old''. An image with attribute \( j \) under tag \( i \) is denoted as \( x_{i,j} \) or \( y_{i,j} \), and similarly for image features.

As discussed in Section~\ref{sec1}, GAN-based methods offer favorable controllability but struggle to align style codes with attribute semantics, while diffusion-based methods generate high-quality images but lack precise attribute control due to the entanglement of semantic directions among different attributes. 
This motivates us to propose AttDiff-GAN, a hybrid framework that combines the strengths of both GANs and diffusion models while addressing their weaknesses to improve facial attribute editing.
However, integrating GANs and diffusion models into a unified framework is challenging due to the inconsistency between GAN's one-step adversarial training and diffusion's multi-step denoising process. To overcome this, we decouple facial attribute editing into feature-level adversarial learning with a generator and discriminator for attribute editing, and conditional denoising for image generation, as shown in Fig.~\ref{fig_2}. Our generator incorporates PriorMapper $M$ and RefineExtractor $E$ to improve style-attribute alignment, while the AttributeEditor $F$, conditioned on style codes, learns flexible, controllable attribute editing under the supervision of Discriminator $D$.
Unlike diffusion-based methods that rely on semantic directions, our approach performs conditional denoising based on the edited semantic codes from the generator, enabling precise editing.
Concretely, taking age editing from young to old ($j \rightarrow j'$) as an example, the target style code $s_{i,j'}$ is first obtained from either $M$ or $E$. Then, under the supervision of the discriminator $D$, the AttributeEditor $F$ edits the input image $x_{i,j}$ conditioned on $s_{i,j'}$, producing a semantic code $c_{i,j'}$. Finally, the denoising network $\epsilon$ iteratively refines the initial latent noise $x_{T,(i,j)}$ corresponding to $x_{i,j}$, conditioned on $c_{i,j'}$, to generate the final edited image $x_{i,j'}$.

In the following sections, we  provide a detailed description of each component of AttDiff-GAN and present the corresponding training objectives.

\subsection{PriorMapper and RefineExtractor}
\label{sec3.2}
Our approach flexibly generates style codes through either latent-guided or reference-guided manner, implemented by the following two modules:

\noindent \textbf{PriorMapper.} 
In the latent-guided manner, existing methods\cite{DalvaTPAMI2023, LiCVPR2021,HuangAAAI2024, ren2025facial} use a Mapper implemented by an MLP network to directly map randomly sampled noise $z$ to style codes, without considering facial features. This lack of facial priors makes it difficult for the generated style codes to align accurately with the intended attributes. As a result, in cases involving complex facial features, the style code either fails to correctly manipulate the target attribute (e.g., “-Smile” in Fig.~\ref{fig_4-1} with VecGAN++, SDGAN, and BSGAN using the MLP-based Mapper) or causes unintended modifications (e.g., ``+Male'' in Fig.~\ref{fig_4-1} with the same methods).

To overcome this limitation, we propose PriorMapper, which uses  transformer to incorporate the facial prior of the image to be edited into the style code generation, improving alignment with the target attributes.  
Specifically, the process of generating style codes with PriorMapper is shown in Fig.~\ref{fig_2} (a). First, a sequence \( z \) sampled from a Gaussian distribution is mapped into a  token sequence through $Token$ $Embed$. Next, the input image is processed by a CNN to extract facial prior features, which, along with the token sequence, is fed into the $Transformer$ $Encoder$ for cross-attention interaction.  Finally, the  output sequence from the $Transformer$ $Encoder$ is concatenated and passed through $Linear$ layer to generate the style code \( s_{i,j'} \). To minimize interference between different attributes, the $Transformer$ $Encoder$ is indexed by tag \( i \), while the subsequent $Linear$ layer is indexed by attribute \( j' \). Formally, given the randomly sampled sequence \( z \), the input image \( x_{i,j} \), and the target attribute \( j' \) under tag $i$, PriorMapper generates style code $s_{i,j'} = M_{i,j'}(z,x_{i,j})$.

\noindent \textbf{RefineExtractor.} 
Unlike generating style codes from noise, reference-guided manner extract style codes from the provided reference image. Precise extraction depends on the extractor's understanding of global facial semantics. However, existing methods\cite{DalvaTPAMI2023, LiCVPR2021,HuangAAAI2024, ren2025facial} rely solely on CNNs, which have limited capacity for capturing global context, leading to suboptimal editing results, as shown in Fig.~\ref{fig_4-2}. 

To address this, we propose RefineExtractor, which combines the efficiency of CNNs with the global context modeling capability of Transformer. The process of extracting style codes with RefineExtractor is shown in Fig.~\ref{fig_2} (b). First, the reference image is processed by a CNN to extract intermediate features, which are then patched into a token sequence. Next, a learnable token is added to the sequence and passed into the $Transformer$ $Encoder$ for cross-attention interaction. Finally, the learnable token, after interaction, is passed through $Linear$ layer to generate the style code $s_{i,j'}$.  Particularly, tag  $i$ is used to retrieve the learnable token and the final linear layer. Formally, given a reference image $y_{i,j'}$ and tag $i$, $E$ extracts the style code $s_{i,j'} = E_i(y_{i,j'})$.

\begin{algorithm}[t]
\caption{AttDiff-GAN}\label{Alg1}
\KwIn{PriorMapper $M$, RefineExtractor $E$, AttributeEditor $F$, Denoising Unet $\epsilon_{\theta}$, source image $x_{i,j}$, reference image $y_{i,j'}$}
\KwOut{editing image $x_{i,j'}$}
\nlset{1}$s_{i,j} = E_i(x_{i,j})$
\footnotesize
\tcp*[f]{\color{gray} Inversion:1-7}\\
\normalsize
\nlset{2}$c_{i,j} = F_{i}(x_{i,j},s_{i,j})$\\
\nlset{3}$x_{0,(i,j)} \leftarrow x_{i,j}$\\
\SetInd{0.5em}{0.3em}
\nlset{4}\textbf{for} $t = 1, \dots, T$ \textbf{do}\\

\normalsize
\nlset{5}\footnotesize
\hspace{1em}$x_{t,(i,j)}\! =\! \sqrt{\alpha_{t}} \left( \frac{x_{t-1,(i,j)}\! -\! \sqrt{1 - \alpha_{t-1}} \, \epsilon_\theta(x_{t-1,(i,j)}, t\!-\!1,c_{i,j})}{\sqrt{\alpha_{t-1}}} \right) + $\\
 $\hspace{7.6em}\sqrt{1 - \alpha_{t}} \, \epsilon_\theta(x_{t-1,(i,j)}, t-1,c_{i,j}) $

\normalsize
\nlset{6}\textbf{end for}\\
\nlset{7} \textbf{switch} (Guided Manner):
\footnotesize
\tcp*[f]{\color{gray} Generation:7-17}\\
\normalsize
\nlset{8} \hspace{1.6em}\textbf{case 1:} Reference-guided \\
\nlset{9}\hspace{5em}$s_{i,j'} = E_i(y_{i,j'})$\\
\nlset{10}\hspace{1.6em}\textbf{case 2:} Latent-guided \\
\nlset{11}\hspace{5em}$s_{i,j'}=M_{i,j'}(z,x_{i,j}),$ $z\sim\mathcal{N}(0, \mathbf{I})$\\
\nlset{12}\textbf{end switch}

\nlset{13}$c_{i,j'} = F_{i}(x_{i,j},s_{i,j'})$\\

\SetInd{0.5em}{0.3em}
\nlset{14}\textbf{for} $t = T, \dots, 1$ \textbf{do}
\footnotesize\\
\normalsize
\nlset{15}\footnotesize\hspace{1em}$x_{t-1,(i,j')}=\sqrt{\alpha_{t-1}} \left( \frac{x_{t,(i,j')} - \sqrt{1 - \alpha_t}\epsilon_\theta(x_{t,(i,j')}, t,c_{i,j'})}{\sqrt{\alpha_t}} \right) +$\\ $\hspace{8em} \sqrt{1 - \alpha_{t-1}}\epsilon_\theta(x_{t,(i,j')}, t,c_{i,j'})$\\
\normalsize
\nlset{16}\textbf{end for}\\
\nlset{17}    $x_{i,j'} \leftarrow x_{0,(i,j')}$\\ 
\nlset{18}\Return $x_{i,j'}$ 

\end{algorithm}

\subsection{AttributeEditor}\label{sec3.3}
Existing diffusion-based methods~\cite{PreechakulCVPR2022, KimCVPR2023} typically achieve facial attribute editing by operating latent codes along semantic directions. However, semantic directions between different attributes are often not orthogonal\cite{shen2020interpreting}, which makes it easy to unintentionally alter other unrelated attributes or content when modifying the target attribute (e.g.,``+Young'' and ``-Young'' in Fig.~\ref{fig_4-1} with HFGI, StyleRes, and DiffAE using semantic directions). To address this issue, we introduce a dedicated editing network, AttributeEditor, to replace the semantic directions for attribute editing. As shown in Fig.~\ref{fig_2} (c), $F$ consists of four components: $Input$ $Blocks$, $Translator$, $Middle$ $Blocks$, and $Output$ $Layer$, where the $Translator$ follows the design of \cite{LiCVPR2021}, and the remaining modules are based on the semantic encoder from \cite{PreechakulCVPR2022}. 
Specifically, in $F$, the $Input$ $Blocks$ takes the image $x_{i,j}$ to be edited as input and extracts facial intermediate features $f_{i,j}$ through a series of downsampling blocks. The $Translator$ consists of a series of AdaIN~\cite{HuangICCV2017} blocks, which edit the attribute of feature $f_{i,j}$ from $j$ to $j'$ based on the style code $s_{i,j'}$, resulting in $f_{i,j'}$. 
Finally, $f_{i,j'}$ is fed into the remaining layers of $F$ to obtain the semantic code $c_{i,j'}$. Formally, given the image $x_{i,j}$ to be edited, tag $i$, and the style code $s_{i,j'}$, $F$ outputs the semantic code $ c_{i,j'} = F_i(x_{i,j}, s_{i,j'}) $.

\subsection{Discriminator}\label{sec3.4}
We introduce a discriminator to ensure that input images are both correctly modified and indistinguishable from real images. However, diffusion models require multi-step denoising to generate clean images, which is fundamentally at odds with the one-step adversarial learning paradigm of GANs. Forcefully applying adversarial supervision throughout the entire denoising process results in prohibitive computational cost and leads to unstable optimization. To address this, we isolate adversarial learning from the denoising process by shifting it from pixel space to feature space.
Specifically, we use the unedited feature $f_{i,j}$ as the real sample and the edited feature $f_{i,j'}$ as the fake sample to train the discriminator. As shown in Fig.~\ref{fig_2} (d), the feature discriminator consists of convolutional layer $Conv$, $Down$ $Blocks$ (LReLU-Conv-AvgPool-LReLU-Conv), and  $Linear$ output layers. The output layer adopts a multi-branch design indexed by tag $i$ and attribute $j$, where each branch is responsible for distinguishing whether the corresponding attribute feature is real or fake (edited).
This design encourages realistic feature generation while ensuring correct attribute editing. For example, when $j$ denotes ``young'' and $j'$ denotes ``old'', $f_{i,j}$ and $f_{i,j'}$ are fed into different branches as real and fake samples, respectively. The ``old'' branch learns to distinguish real aged features from edited ones, while the $Translator$ is required to correctly transform ``young'' into ``old'' and maintain realism to fool the discriminator. The ``young'' branch follows the same principle.

\subsection{Denoising UNet}\label{sec3.5}
After obtaining the semantic code, the edited image is generated through an iterative denoising process. In this work, we adopt conditional DDIM (cDDIM)\cite{PreechakulCVPR2022} for this process because of its strong invertibility in both the noising and denoising stages.
Specifically, given an input image $x_{i,j}$, we first obtain its initial latent noise $x_{T,(i,j)}$ via cDDIM inversion using the denoising UNet $\epsilon_\theta$. Starting from $x_{0,(i,j)}= x_{i,j}$, $x_{T,(i,j)}$  is  computed by iteratively applying:
\begin{equation}\label{eq2}
\small
\begin{aligned}
x_{t,(i,j)}\! = \!\!&\  \sqrt{\alpha_{t}}\!\! \left(\! \frac{x_{t-1,(i,j)}\! - \!\sqrt{1\! -\! \alpha_{t-1}} \, \epsilon_\theta(x_{t-1,(i,j)}, t\!-\!1,c_{i,j})}{\sqrt{\alpha_{t-1}}} \!\right) \\
&+\! \sqrt{1\! - \!\alpha_{t}} \, \epsilon_\theta(x_{t-1,(i,j)}, t\!-\!1,c_{i,j}).
\end{aligned}
\end{equation}
This establishes a deterministic mapping between the input image and its latent representation.

During the generative process, the denoising UNet is conditioned on the semantic code $c_{i,j'}$ to perform image denoising starting from $x_{T,(i,j)}$. At each timestep $t$,  UNet predicts the noise component as $\epsilon_\theta(x_{t,(i,j')}, t, c_{i,j'})$, where at $t = T$, the input latent is $x_{T,(i,j)}$, and the latent variable is updated using the cDDIM sampling rule:
\begin{equation}\label{eq2}
\small
\begin{aligned}
x_{t-1,(i,j')}\!=&\sqrt{\alpha_{t-1}}
\left(
\frac{x_{t,(i,j')}\!-\!\sqrt{1-\alpha_t}\,\epsilon_\theta(x_{t,(i,j')},t,c_{i,j'})}
{\sqrt{\alpha_t}}
\right) \\
&\!+\!\sqrt{1-\alpha_{t-1}}\,\epsilon_\theta(x_{t,(i,j')},t,c_{i,j'}).
\end{aligned}
\end{equation}
\normalsize
By conditioning on $c_{i,j'}$ to iteratively update the latent variable, the model progressively transforms the input image toward the target attribute, yielding the edited result $x_{(i,j')} = x_{0,(i,j')}$. The overall inversion and generative process is summarized in Algorithm~\ref{Alg1}.

\subsection{Training Objectives}\label{sec3.6}
To reduce the computational cost of iterative denoising in diffusion models, we formulate the training objectives in feature space. The overall objective consists of reconstruction, adversarial, and classification losses, which are detailed below.

\noindent\textbf{Reconstruction Loss.}
To encourage the edited results to preserve attribute-irrelevant content of the input image, we consider two feature reconstruction cases:
\begin{enumerate}
\item[{1)}]  $f'_{i,j}=T_i(f_{i,j}, s_{i,j})$, where $T$ denotes the Translator in $F$ and $s_{i,j}=E_i(x_{i,j})$. By using the style code $s_{i,j}$ extracted from the input image $x_{i,j}$, the output is expected to reconstruct the input feature.

\item[{2)}] $f''_{i,j}=T_i(f_{i,j'}, s_{i,j})$, where $f_{i,j'}=T_i(f_{i,j},s_{i,j'})$ and $s_{i,j'}=M_{i,j'}(z,x_{i,j})$. We first edit the input feature $f_{i,j}$ to the target attribute $f_{i,j'}$ using the latent-guided style code $s_{i,j'}$, and then reconstruct the original feature $f_{i,j}$ using the reference-guided style code $s_{i,j}$.
\end{enumerate}
Based on these cases, the reconstruction loss is defined as:
\begin{equation}\label{eq3}
\begin{aligned}
{\mathcal{L}_{rec}=||f'_{i,j}-f_{i,j}||_1+||f''_{i,j}-f_{i,j}||_1}.
\end{aligned}
\end{equation}

\noindent\textbf{Adversarial Loss.} As described in Section~\ref{sec3.4}, we introduce adversarial training in feature space to ensure that the features are both correctly modified and indistinguishable from real ones. Formally, the adversarial loss is defined as:
\begin{equation}\label{eq4}
\begin{aligned}
\mathcal{L}_{adv}=&2\log D_{i,j}(f_{i,j})+\log(1-D_{i,j'}(f_{i,j'}))\\
&+\log(1-D_{i,j}(f''_{i,j})).
\end{aligned}
\end{equation}
This objective encourages the discriminator to accurately distinguish real from fake features. Simultaneously, it guides the generator (i.e., \(M\), \(E\), and \(F\)) by penalizing both incorrectly edited features and reconstruction artifacts, thereby helping it produce edited features that are indistinguishable from real ones under the target attribute branch, and reconstructed features remain consistent with the original attribute.

\noindent\textbf{Classification Loss.} We introduce a classification loss on the semantic code $c_{i,j'}$ to further ensure that the target attribute is correctly reflected in the edited result:
\begin{equation}\label{eq5}
\begin{small}
\begin{aligned}
\mathcal{L}_{cls}=-l^\top\log{}C(c_{i,j'})-(1-l^\top)\log{}(1-C(c_{i,j'})),
\end{aligned}
\end{small}
\end{equation}
where $l\in \mathbb{R}^{n}$ denotes the  binary target attribute vector, $\top$ indicates the transpose operation, and $C(\cdot)\in\mathbb{R}^{n}$ outputs the predicted probabilities over $n$ attributes.

\noindent\textbf{Full Loss.} The overall objective is formulated as:
\begin{equation}\label{eq6}
\begin{aligned}
\mathop{\min}\limits_{M,E,F}\mathop{\max}\limits_{D} (\mathcal{L}_{rec}+\mathcal{L}_{adv}+\mathcal{L}_{cls}{).}
\end{aligned}
\end{equation}

\begin{table}[t]
\caption{Comparison of Our Method and State-of-the-Art in Editing strategies and Implementation Frameworks}
\label{Table 0}
\centering
\makegapedcells
\setcellgapes{2pt}
\setlength\tabcolsep{1pt}
\small
\begin{tabular}{!{\vrule width 0.8pt}l!{\vrule width 0.8pt}c|c|c|c|c!{\vrule width 0.8pt}c|c!{\vrule width 0.8pt}}
\toprule
\multirow{3}{*}{Methods} 
& \multicolumn{5}{c!{\vrule width 0.8pt}}{Editing strategies} 
& \multicolumn{2}{c!{\vrule width 0.8pt}}{Frameworks} \\
\cline{2-8}
& \multicolumn{1}{c|}{Attr.} 
& \multicolumn{1}{c|}{Sem.} 
& \multicolumn{1}{c|}{Lat.-} 
& \multicolumn{1}{c|}{Ref.-} 
& \multicolumn{1}{c!{\vrule width 0.8pt}}{Text-} 
& \multirow{2}{*}{GANs} 
& \multirow{2}{*}{DMs} \\
& vec. & dir. & gui. & gui. & gui. & & \\
\midrule
HFGI {\scriptsize (CVPR'22)}\cite{WangCVPR2022} &  & \ding{51} &  &  &  & \ding{51} &  \\
StyleRes {\scriptsize (CVPR'23)}\cite{PehlivanCVPR2023} &  & \ding{51} &  &  &  & \ding{51} &  \\
VecGAN{\scriptsize ++} {\scriptsize (TPAMI'23)}\cite{DalvaTPAMI2023} &  &  & \ding{51} & \ding{51} &  & \ding{51} &  \\
InterGAN {\scriptsize (TCSVT'24)}\cite{HuangTCSVT2024} & \ding{51} &  &  &  &  & \ding{51} &  \\
SDGAN {\scriptsize (AAAI'24)}\cite{HuangAAAI2024} &  &  & \ding{51} & \ding{51} &  & \ding{51} &  \\
BSGAN {\scriptsize (ESWA'25)}\cite{ren2025facial} &  &  & \ding{51} & \ding{51} &  & \ding{51} &  \\
DiffAE {\scriptsize (CVPR'22)}\cite{PreechakulCVPR2022} &  & \ding{51} &  &  &  &  & \ding{51} \\
IP2P {\scriptsize (CVPR'23)}\cite{brooks2023instructpix2pix} &  &  &  &  & \ding{51} &  & \ding{51} \\
I-CLIP {\scriptsize (CVPR'25)}\cite{chen2025instruct} &  &  &  &  & \ding{51} &  & \ding{51} \\
AttDiff-GAN &  &  & \ding{51} & \ding{51} &  & \ding{51} & \ding{51} \\
\bottomrule
\end{tabular}
\end{table}

\begin{figure*}[!ht]
\centering
\includegraphics[width=7.1in]{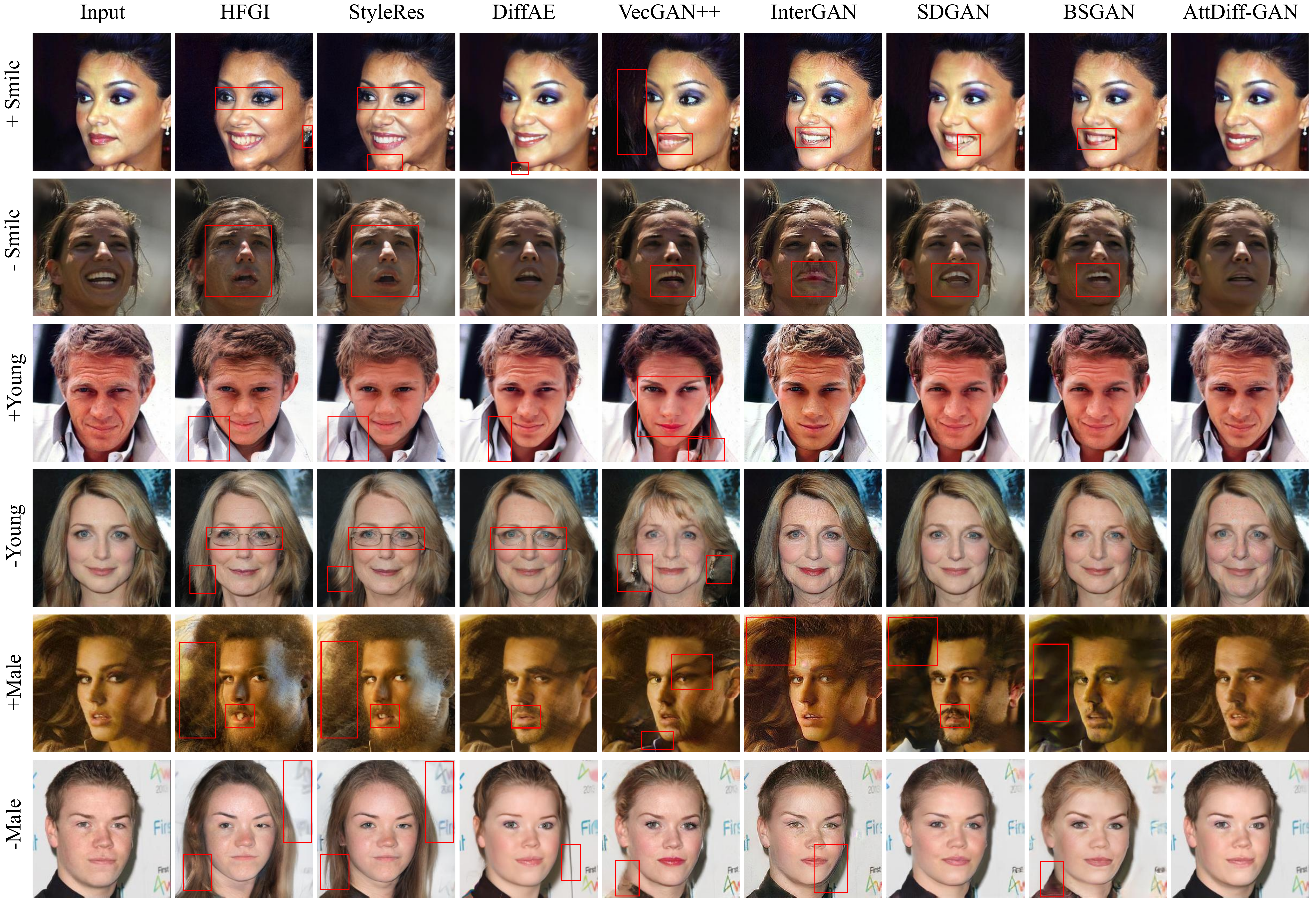}
\caption{Qualitative results of latent-guided evaluation. In the figures, we highlight the problematic areas in the generated images that require magnification to observe the detailed differences.  }
\label{fig_4-1}
\end{figure*}

\section{Experiments}\label{sec4}

\subsection{Database and Baselines}\label{sec4.1}
Consistent with previous work\cite{LiCVPR2021,DalvaTPAMI2023,HuangAAAI2024}, we conducted our experiments using the CelebA-HQ dataset~\cite{KarrasICLR2018}, which contains 30,000 high-resolution facial images annotated with binary attributes. Following the methodology of~\cite{DalvaTPAMI2023,yang2021l2m}, we partitioned the dataset into 27,176 images for training and 2,824 images for testing. For evaluation, we focus on three widely used facial attributes: \textit{Smile}, \textit{Young}, and \textit{Male}. To provide a comprehensive comparison, we examine seven state-of-the-art face attribute editing techniques, representing four primary editing strategies: attribute vectors, semantic directions, latent-guided, and reference-guided approaches. These methods are detailed in Table~\ref{Table 0}.

Beyond conventional face-attribute editing techniques, we broaden our comparison to include the latest advancements in text-to-image foundation models. These models, trained on large-scale image-text datasets, exhibit remarkable flexibility in handling a wide range of tasks. In particular, we focus on IP2P~\cite{brooks2023instructpix2pix} and I-CLIP~\cite{chen2025instruct}, which excel at generating precise image modifications based on textual descriptions. Their ability to preserve irrelevant content during the editing process makes them especially well-suited for face attribute editing tasks.

\subsection{Training Details and Evaluation Metrics}\label{sec4.2}

\vspace{0.5em}\noindent {\textbf{Training Details.} 
In our model, the denoising UNet $\epsilon_\theta$ and  the AttributeEditor $F$ (excluding the Translator $T$)  are initialized with pre-trained weights from DiffAE~\cite{PreechakulCVPR2022}. These components are kept frozen during training to preserve their generative and semantic modeling capabilities. The remaining modules, including the PriorMapper $M$, RefineExtractor $E$, Translator $T$, and the feature discriminator $D$, are trained jointly.
We adopt Adam optimizer for all trainable components. The learning rates are set to $1\times10^{-4}$ for $E$, $T$, and $D$, and $1\times10^{-6}$ for $M$. 
All input images are resized to $256\times256$, and the batch size is set to 16. The model is trained for 60 epochs. Source code for AttDiff-GAN is publicly available online\footnote{https://github.com/WeMiHuang/AttDiff-GAN}.}

\vspace{0.5em}\noindent  \textbf{Evaluation Metrics.}
In face attribute editing, two key metrics are the quality of the edited images and the editing success rate. In line with previous methods, we employ the Fréchet Inception Distance (\textbf{FID})~\cite{HeuselNIPS2017} to measure the statistical disparity between edited and authentic images, which helps assess the realism of the generated images. To evaluate the editing success rate, we use the commonly accepted metric of attribute editing accuracy (\textbf{Acc})~\cite{DalvaTPAMI2023, HuangAAAI2024}. This metric relies on an attribute classifier built on ResNet-18~\cite{HeCVPR2016}, which is trained using the CelebA-HQ dataset and achieves a $95.0\%$ accuracy on the test set.

\begin{table*}[!ht]
\caption{Quantitative results of latent-guided evaluation}\label{Table 4-1}
      \centering
      \makegapedcells
      \setcellgapes{2pt}
      \setlength\tabcolsep{6pt}
      \begin{tabular}{c || c  c | c c| c  c | c c|c c|c c|c c}
        \hline
        \multirow{2}{*}{Method} &\multicolumn{2}{c|}{+ Smile}&\multicolumn{2}{c|}{- Smile}& \multicolumn{2}{c|}{+ Young}& \multicolumn{2}{c|}{- Young}& \multicolumn{2}{c|}{+ Male}& \multicolumn{2}{c|}{- Male}&\multicolumn{2}{c}{Average}\\
        \cline{2-15}
        &$\downarrow{}$FID &$\uparrow{}$Acc &$\downarrow{}$FID &$\uparrow{}$Acc &$\downarrow{}$FID &$\uparrow{}$Acc &$\downarrow{}$FID &$\uparrow{}$Acc&$\downarrow{}$FID &$\uparrow{}$Acc&$\downarrow{}$FID &$\uparrow{}$Acc&$\downarrow{}$FID &$\uparrow{}$Acc\\
        \hline
        \hline
         HFGI\cite{WangCVPR2022}&29.59&83.77&36.69&91.23&54.37&80.39&51.81&81.34&67.33&97.37&52.17&88.68&48.66&87.13\\
         \hline
         StyleRes\cite{PehlivanCVPR2023}&23.05&94.61&25.81&96.71&50.99&87.92&32.36&82.84&63.38&92.13&49.20&94.57&40.79&91.46\\
         \hline
         DiffAE\cite{PreechakulCVPR2022}&20.42&94.21&25.50&90.86&47.43&78.59&26.27&92.22&42.98&96.76&41.77&97.01&34.06&91.60\\
         \hline
         VecGAN++\cite{DalvaTPAMI2023}&19.91&95.83&25.66&90.56&40.65&96.37&26.83&82.50&39.43&99.81&34.00&90.31&31.08&92.56\\
         \hline
         SDGAN~\cite{HuangAAAI2024}&21.80&88.38&26.74&86.85&40.51&82.56&32.23&74.86&45.89&93.65&34.37&95.65&33.59&86.99\\
         \hline
         InterGAN~\cite{HuangTCSVT2024}&20.51&95.90&25.61&96.19&44.12&92.99&28.45&\textbf{92.63}&41.16&93.05&35.64&93.82&32.58&94.09\\
         \hline
         BSGAN\cite{ren2025facial}&19.31&80.55&26.22&89.43&42.23&80.96&33.69&75.05&44.79&85.11&33.79&89.61&33.34&83.45\\
         \hline
         AttDiff-GAN &\textbf{17.61}&\textbf{99.48}&\textbf{25.48}&\textbf{97.38}&\textbf{37.12}&\textbf{96.40}&\textbf{26.15}&86.58&\textbf{28.50}&\textbf{100}&\textbf{32.93}&\textbf{99.54}&\textbf{27.96}&\textbf{96.56}\\
        \hline
      \end{tabular}
\end{table*}

\begin{table*}[!ht]
\caption{Quantitative results of reference-guided evaluation}\label{Table 4-2}
      \centering
      \makegapedcells
      \setcellgapes{2pt}
      \setlength\tabcolsep{6pt}
      \begin{tabular}{c || c  c | c c| c  c | c c|c c|c c|c c}
        \hline
        \multirow{2}{*}{Method} &\multicolumn{2}{c|}{+ Smile}&\multicolumn{2}{c|}{- Smile}& \multicolumn{2}{c|}{+ Young}& \multicolumn{2}{c|}{- Young}& \multicolumn{2}{c|}{+ Male}& \multicolumn{2}{c|}{- Male}&\multicolumn{2}{c}{Average}\\
        \cline{2-15}
        &$\downarrow{}$FID &$\uparrow{}$Acc &$\downarrow{}$FID &$\uparrow{}$Acc &$\downarrow{}$FID &$\uparrow{}$Acc &$\downarrow{}$FID &$\uparrow{}$Acc&$\downarrow{}$FID &$\uparrow{}$Acc&$\downarrow{}$FID &$\uparrow{}$Acc&$\downarrow{}$FID &$\uparrow{}$Acc\\
        \hline
        \hline
         VecGAN++\cite{DalvaTPAMI2023}&20.60&97.88&26.13&95.11&36.86&93.44&26.61&80.38&46.97&86.95&33.68&85.97&31.80&89.95\\
        \hline
         SDGAN~\cite{HuangAAAI2024}&21.02&97.49&27.98&97.47&40.94&90.76&32.75&\textbf{87.85}&35.71&95.06&35.67&93.75&32.34&93.73\\
         \hline
         BSGAN\cite{ren2025facial}&19.37&85.43&25.84&89.27&40.69&87.80&32.93&86.17&37.19&92.93&33.49&95.20&31.58&89.46\\
         \hline
         AttDiff-GAN &\textbf{17.76}&\textbf{99.16}&\textbf{25.56}&\textbf{97.95}&\textbf{36.17}&\textbf{94.96}&\textbf{26.22}&87.18&\textbf{28.19}&\textbf{99.87}&\textbf{33.29}&\textbf{99.09}&\textbf{28.03}&\textbf{96.36}\\
        \hline
      \end{tabular}
\end{table*}

\begin{figure*}[!ht]
\centering
\includegraphics[width=6.4in]{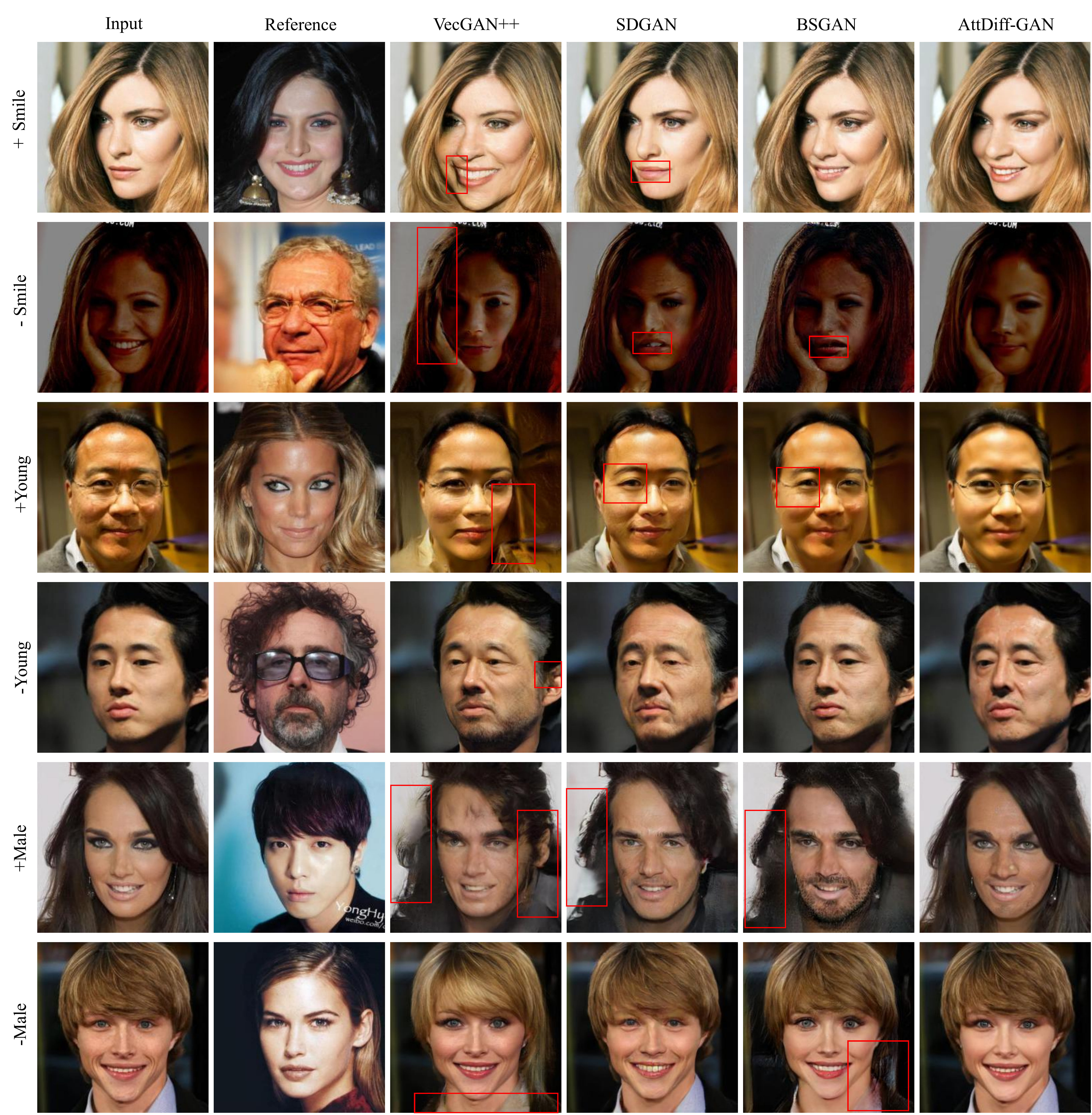}
\caption{Qualitative results of reference-guided evaluation.}
\label{fig_4-2}
\end{figure*}

\subsection{Comparison with Face Attribute Editing Baselines}\label{sec4.3}
In this section, we evaluate the proposed method by comparing it with recent face attribute editing approaches, using both qualitative and quantitative analyses from the latent-guided and reference-guided perspectives.

\subsubsection{Latent-Guided Evaluation}\label{sec4.3.1}
In this part, we focus on evaluating the performance of our method in latent-guided attribute editing. The comparison includes latent-guided editing methods, as well as attribute vector- and semantic direction-based methods, which similarly do not require additional reference images. The qualitative results are presented in Fig.~\ref{fig_4-1}, from which we draw the following observations:

\begin{itemize}
    \item A common limitation of the compared methods is their tendency to unintentionally alter unrelated regions during attribute editing, such as eye gaze, background, earrings, hair, clothing, and mouth shape, as highlighted by the red boxes in the figure.

    \item Regarding attribute editing, when ``-Smile'', HFGI and StyleRes incorrectly alter gender, while VecGAN++, InterGAN, SDGAN, and BSGAN fail to completely remove the smile. When ``-Young'', HFGI, StyleRes, and DiffAE mistakenly add eyeglasses.

    \item Some methods introduce artifacts that compromise realism. For example, VecGAN++, InterGAN, SDGAN, and BSGAN generate blurry and unrealistic teeth when ``+Smile'', and distorted mouth features when ``-Smile''. HFGI and StyleRes produce outputs with noticeable blur and noise when ``+Male''.

    \item In contrast to the challenges faced by these methods, our approach demonstrates superior performance by effectively preserving unrelated content, accurately editing the target attributes, and generating highly realistic results.
\end{itemize}

The quantitative results, as shown in Table~\ref{Table 4-1}, demonstrate that our method consistently outperforms existing techniques across nearly all attribute editing tasks. Specifically, it achieves an average FID of $27.96$, representing a $3.12$ improvement over the leading method, VecGAN++ (31.08). Additionally, our method achieves superior accuracy with a score of $96.56\%$, surpassing the previous best performer, InterGAN ($94.09\%$), by $2.47$ percentage points. These results demonstrate the significant advancements our approach makes in delivering high-quality attribute editing results.

\subsubsection{Reference-Guided Evaluation}
In this evaluation, we focus exclusively on comparing methods for reference-guided editing. The qualitative comparison results are shown in Fig.~\ref{fig_4-2}. Similar to the latent-guided evaluation, the compared methods still fail to preserve irrelevant content, such as eyeglasses, hair, and clothing. Additionally, VecGAN++ generates unrealistic mouth corners when adding a smile, alters gender attributes and introduces facial distortions when attempting to add young. SDGAN struggles with adding a smile, while both SDGAN and BSGAN introduce mouth distortions when removing a smile, affecting realism. In contrast, our method achieves the best editing results in all comparisons.

\begin{figure*}[!ht]
\centering
\includegraphics[width=6.4in]{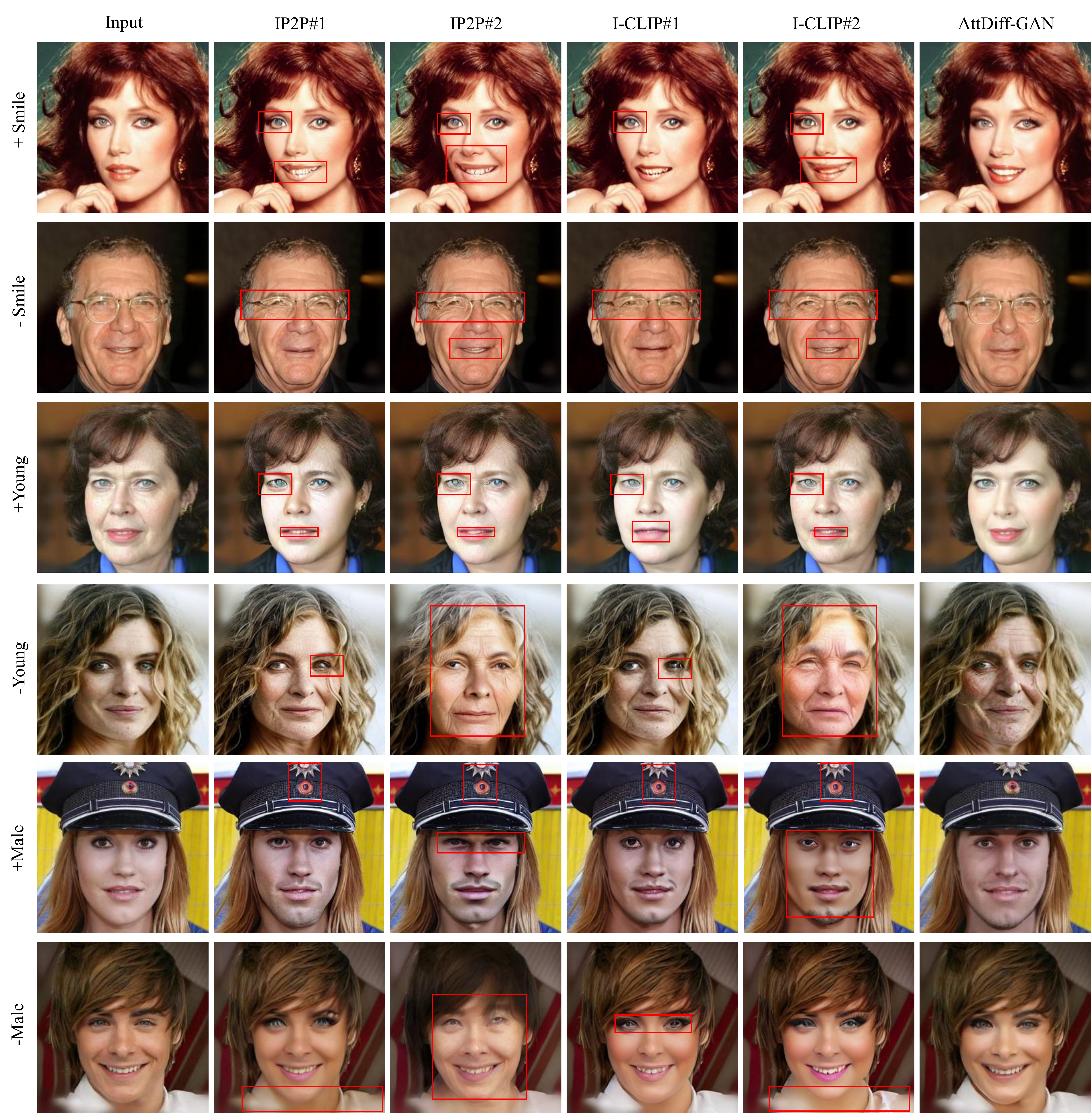}
\caption{Comparison between our method and text-to-image models.}
\label{fig_4-3}
\end{figure*}

As shown in Table~\ref{Table 4-2}, the quantitative results demonstrate the consistent superiority of our method over existing approaches in nearly all attribute editing tasks. Our method achieves an average FID of $28.03$, improving upon the state-of-the-art BSGAN (31.58) by $3.55$. Additionally, it reaches the highest mean accuracy $96.36\%$, surpassing SDGAN ($93.73\%$) by $2.63$ percentage points. These findings demonstrate the effectiveness of our method in enhancing both quality and editing success rate.

\subsection{Comparison with Text-to-Image Models}\label{sec4.4}
Recent large-scale text-to-image models, such as Stable Diffusion\cite{rombach2022high} and Imagen\cite{saharia2022photorealistic}, have made significant strides in generating detailed images from text. However, these models often lack precise editing abilities, where small text changes lead to significant output variations. To address these issues, instruction-driven editing methods, trained on image pairs and corresponding instructions, have emerged, offering refined control for localized edits. Given the need to preserve irrelevant regions during editing, we explore two instruction-driven models as baselines: IP2P~\cite{brooks2023instructpix2pix} and I-CLIP~\cite{chen2025instruct}.

Both IP2P and I-CLIP's desirable editing results depend on the appropriate text instructions and the specific guiding scales $s_T$ and $s_I$, and are influenced by random sampling $x_T$. Adjusting $s_I$ makes the output closer to the original, while $s_T$ controls the intensity of the edit. For a fair comparison, we adjust the  text instructions, $s_T$, $s_I$, and $x_T$ for each image to achieve optimal results, labeled as IP2P\#1 and I-CLIP\#1. To assess stability conservatively, we fix the instructions and scales, varying only $x_T$, resulting in IP2P\#2 and I-CLIP\#2. The experimental results shown in Fig. 5 reveal the following:

\begin{itemize}
\item The editing results of IP2P\#1 and I-CLIP\#1 struggle to preserve fine details. For example, when applying ``+Smile'' and ``+Young'', they result in unrealistic textures in the eyes and teeth. Furthermore, when using ``+Smile'', ``+Male'', and ``-Male'', they fail to preserve irrelevant elements such as eyeglasses, hats, and clothing.

\item A comparison between IP2P\#1 vs. IP2P\#2 and I-CLIP\#1 vs. I-CLIP\#2 reveals notable instability, even when only $x_T$ is varied. Failures occur when applying ``-Smile'', and noticeable image quality degradation is observed when applying ``-Young'' or ``+Male''. This instability increases time costs in real-world applications.

\item In contrast, our method consistently produces high-quality, realistic edits that remain faithful to the original input. Importantly, this is achieved without the need for extensive parameter adjustments, ensuring greater stability and efficiency in the editing process.

\end{itemize}

Finally, we highlight that, beyond $x_T$, IP2P and I-CLIP are highly sensitive to both text instructions and guiding scales. Achieving satisfactory results requires extensive tuning for each image, making large-scale evaluation unfeasible. Thus, we do not provide quantitative results for these methods.

\begin{figure*}[!ht]
\centering
\includegraphics[width=7.1in]{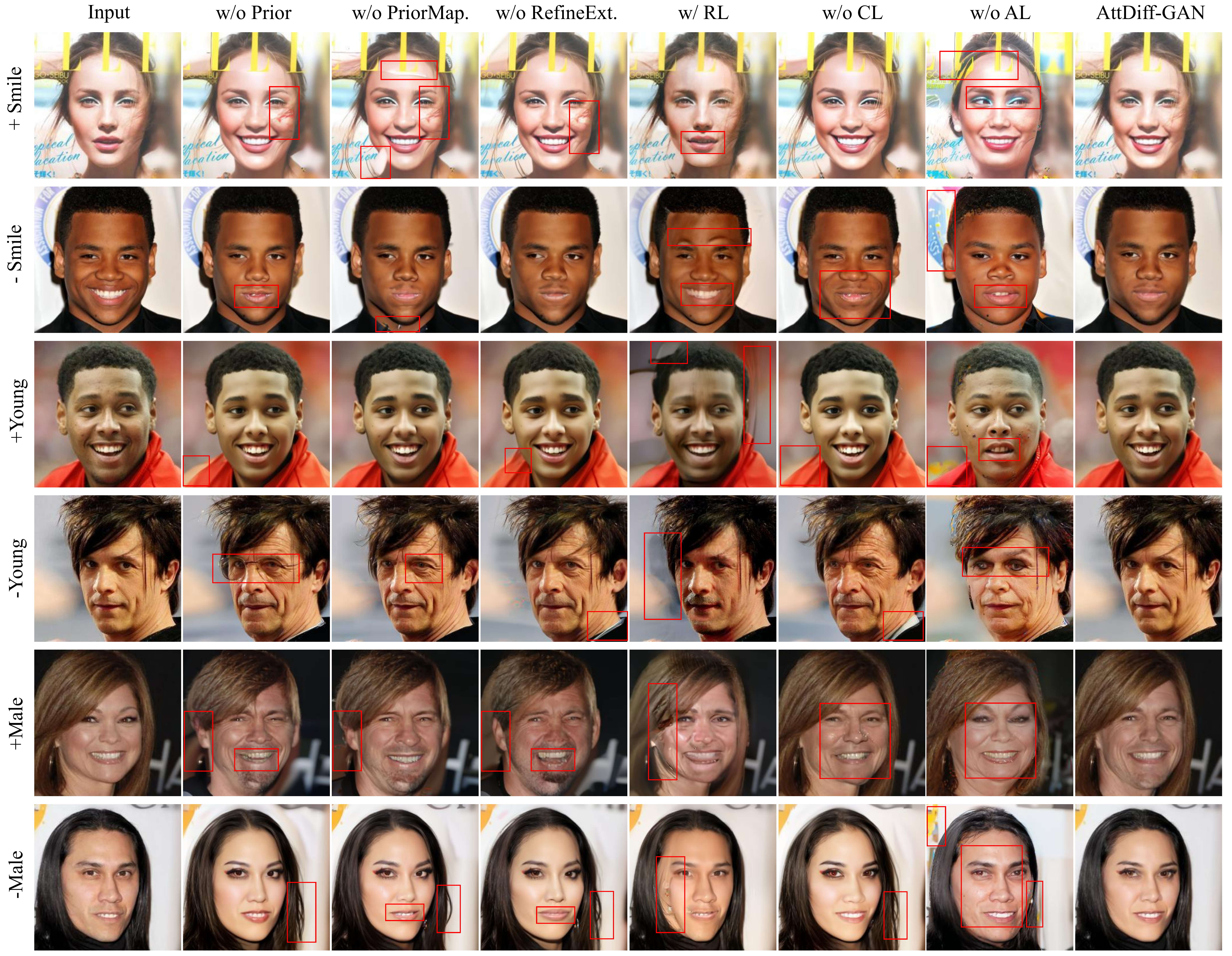}
\caption{Qualitative results of the ablation study.}
\label{fig_4-4}
\end{figure*}

\begin{table}[t]
\caption{Comparison of Editing Time and Model Size Across Methods}
\label{Table_time_param}
\centering
\renewcommand{\arraystretch}{1.2}
\setlength{\tabcolsep}{2.3pt}
\small

\begin{tabular}{
!{\vrule width 0.8pt}l
|c
|c
|c
!{\vrule width 0.8pt}
}
\toprule

\textbf{Methods} & \textbf{Framework} & \textbf{Editing Time (s)} & \textbf{Parameter (M)} \\

\midrule

HFGI\cite{WangCVPR2022}      & \multirow{6}{*}{GANs} & 0.083 & 297.50 \\
StyleRes\cite{PehlivanCVPR2023}  &                        & 0.055 & 390.92 \\
VecGAN++\cite{DalvaTPAMI2023}    &                        & 0.284 & 111.91 \\
InterGAN\cite{HuangTCSVT2024}  &                        & 0.053 & 28.07  \\
SDGAN\cite{HuangAAAI2024}     &                        & 0.030 & 19.67  \\
BSGAN\cite{ren2025facial}     &                        & 0.015& 18.85\\

\midrule

DiffAE\cite{PreechakulCVPR2022}    & \multirow{3}{*}{DMs}  & 3.328 & 168.49 \\
IP2P\cite{brooks2023instructpix2pix}      &                       & 5.752 & 1066.24 \\
I-CLIP\cite{chen2025instruct}    &                       & 5.514 & 1072.61 \\

\midrule

AttDiff-GAN & GANs+DMs            & 4.842 & 726.51 \\

\bottomrule
\end{tabular}
\end{table}

\begin{table*}[h]
\caption{Quantitative results of  ablation study}\label{Table 4-3}
      \centering
      \makegapedcells
      \setcellgapes{2pt}
      \setlength\tabcolsep{6pt}
      \begin{tabular}{c || c  c | c c| c  c | c c|c c|c c|c c}
        \hline
        \multirow{2}{*}{Method} &\multicolumn{2}{c|}{+ Smile}&\multicolumn{2}{c|}{- Smile}& \multicolumn{2}{c|}{+ Young}& \multicolumn{2}{c|}{- Young}& \multicolumn{2}{c|}{+ Male}& \multicolumn{2}{c|}{- Male}&\multicolumn{2}{c}{Average}\\
        \cline{2-15}
        &$\downarrow{}$FID &$\uparrow{}$Acc &$\downarrow{}$FID &$\uparrow{}$Acc &$\downarrow{}$FID &$\uparrow{}$Acc &$\downarrow{}$FID &$\uparrow{}$Acc&$\downarrow{}$FID &$\uparrow{}$Acc&$\downarrow{}$FID &$\uparrow{}$Acc&$\downarrow{}$FID &$\uparrow{}$Acc\\
        \hline
        \hline
         w/o Prior &19.09&98.48&26.17&96.17&37.47&92.36&27.39&83.49&32.61&100&34.59&98.72&29.55&94.87\\

         w/o PriorMap. &18.94&98.55&26.65&95.77&40.36&86.82&27.67&67.61&34.09&99.81&36.99&98.64&30.78&91.22\\

         w/o RefineExt. &18.97&98.74&25.89&96.62&38.62&94.91&26.89&81.94&43.16&100&36.14&99.54&31.61&95.29\\
         \hline
         \hline
         w/o RL &26.86&23.65&31.57&21.11&62.02&17.90&43.06&55.98&90.61&14.08&69.85&32.97&53.99&27.65\\
         
         w/o CL&18.13&98.67&25.90&95.60&39.87&95.47&29.63&83.14&30.37&99.93&32.98&98.81&29.48&95.27\\
         
         w/o AL&25.97&98.13&36.63&96.03&58.62&30.53&57.64&51.54&113.78&5.06&125.63&9.14&69.71&48.40\\
         \hline
         AttDiff-GAN &\textbf{17.61}&\textbf{99.48}&\textbf{25.48}&\textbf{97.38}&\textbf{37.12}&\textbf{96.40}&\textbf{26.15}&\textbf{86.58}&\textbf{28.50}&\textbf{100}&\textbf{32.93}&\textbf{99.54}&\textbf{27.96}&\textbf{96.56}\\
        \hline
      \end{tabular}
\end{table*}

\subsection{Comparison of Editing Efficiency and Model Size}\label{sec4.5}
In this section, we compare the editing time and parameter size of our method with those of state-of-the-art approaches to evaluate both computational efficiency and model complexity.
As shown in Table~\ref{Table_time_param}, along with the image quality and editing accuracy evaluations in Sections~\ref{sec4.3} and~\ref{sec4.4}, we draw the following observations:
\begin{itemize}
    \item Diffusion-based methods incur significantly higher editing time per image than GAN-based methods, primarily due to their reliance on multiple iterative denoising steps for high-quality generation.

    \item Compared to DiffAE, our method is slightly slower but achieves significantly better image quality and editing accuracy. The increase in parameters comes from the transformer-based PriorMapper and RefineExtractor, enabling flexible and  controllable attribute manipulation.

    \item Compared to IP2P and I-CLIP, our method achieves superior editing performance while maintaining faster editing speed and a smaller model size. This demonstrates the effectiveness and efficiency of our approach.
\end{itemize}

\subsection{Ablation Study}\label{sec4.6}
In this work, we design two key networks, PriorMapper and RefineExtractor, and introduce the three essential loss functions that support the hybrid Diffusion-GAN  framework: reconstruction loss, classification loss, and adversarial loss. We validate their effectiveness through ablation experiments, with qualitative and quantitative results shown in Fig.~\ref{fig_4-4} and summarized in Table \ref{Table 4-3}, respectively. Overall, AttDiff-GAN demonstrates the best editing performance in qualitative evaluations and consistently outperforms other ablation variants across all evaluation metrics. Subsequently, we analyze the effectiveness of these networks and loss functions based on the results presented in Fig.~\ref{fig_4-4} and Table \ref{Table 4-3}.

\subsubsection{Analysis on PriorMapper}\label{sec4.5.1}
We remove the input image prior from PriorMapper, denoted as ``w/o Prior'' for brevity.  Qualitatively, compared to AttDiff-GAN, the operation of ``+Smile'' introduces facial distortions, while ``-Smile'' and ``-Young'' result in mouth twisting and incorrect manipulation of eyeglass attributes, respectively. Additionally, the operations of ``+Young'', ``+Male'', and ``-Male'' incorrectly modify the clothing, hair, and mouth shape, respectively. This is primarily due to the network's difficulty in generating style codes that are well-aligned with the target attributes from pure noise in the absence of image priors, which impacts the precision of the editing. Quantitative results further confirm that removing the image prior leads to a decline in both image quality and editing accuracy. For instance, the average FID and accuracy are $1.59$ and $1.69\%$ worse than AttDiff-GAN, respectively.

Furthermore, we replace PriorMapper with the MLP-based Mapper, which is widely used in existing methods\cite{HuangAAAI2024,ren2025facial}, referred to as ``w/o PriorMap.''. Qualitatively, compared to AttDiff-GAN, the operation of ``+Smile'' introduces multiple facial artifacts, ``-Smile'' results in collar contamination, ``-Young'' alters the eye gaze, and both ``+Male'' and ``-Male'' fail to preserve hair details.  Quantitatively, the average FID and accuracy significantly trail behind AttDiff-GAN by 2.82\% and 5.34\%, respectively. These results clearly highlight the effectiveness of the CNN-Transformer hybrid architecture introduced in PriorMapper in improving both image quality and attribute editing accuracy.

\subsubsection{Analysis on  RefineExtractor}\label{sec4.5.2}
We replace RefineExtractor with a CNN-based Extractor, commonly used in existing methods\cite{HuangAAAI2024,ren2025facial}, denoted as ``w/o RefineExt.'' Qualitative results reveal that the operation of ``+Smile'' introduces facial artifacts and contaminates the background. Similarly, the operations of ``+Young'' and ``-Young'' fail to preserve clothing details, while ``+Male'' and ``-Male'' alter the mouth shape and compromise hair details. These results highlight that the introduction of cross-attention mechanisms in the extractor significantly improves feature consistency in the edited images. Quantitative results further confirm that replacing RefineExtractor leads to a noticeable decline in image quality, with a 3.65 increase in the average FID.

\subsubsection{Analysis on Reconstruction Loss}\label{sec4.5.3}
We remove the reconstruction  loss, denoted as ``w/o RL''. Qualitatively, removing the reconstruction loss leads to failures in all attribute editing operations except for ``-Young'', accompanied by blurry images and noticeable artifacts in both the face and background. Quantitatively, we observe a 26.03 increase in average FID and a significant 68.91\% drop in average accuracy. A possible explanation for the decline in accuracy is that,  without the reconstruction  loss constraint on the image content, the model may resort to a strategy similar to adversarial attacks, deceiving the classification losses. The resulting artifacts in the edited images could be a consequence of this behavior.

\subsubsection{Analysis on Classification Loss}\label{sec4.5.4}
We remove the classification loss, denoted as ``w/o CL''. Qualitatively, compared to AttDiff-GAN, the operation of ``-Smile'' results in unnatural mouth texture and nasolabial folds. The operations of ``+Youth'', ``-Youth'', and ``-Male'' fail to preserve clothing and hair details. Furthermore, the operation of ``+Male'' does not fully transform the female to male. Quantitatively, we observe a decline across all editing task metrics, demonstrating that the classification loss is effective in enhancing the model's attribute editing capability.

\subsubsection{Analysis on Adversarial Loss}\label{sec4.5.5}
We remove the adversarial loss, denoted as ``w/o AL''. Qualitative results show that the operation of ``+Smile'' and ``-Young'' introduce foreground font artifacts and eye distortions, respectively. When manipulating ``-Smile'', the model's ability to edit the mouth is weakened, and the operations of ``+Male'' and ``-Male'' result in editing failures. Quantitative results further indicate that removing the adversarial loss has a minimal impact on editing local attributes, such as ``+/-Smile'', but significantly affects global attributes, especially ``+/-Male''. This leads to a decline in both the average FID and accuracy, with worsened values of 41.75 and 48.16\%, respectively.

\section{Conclusion}\label{sec5}
In this paper, we propose AttDiff-GAN, a novel hybrid framework that combines the controllability of GANs with the high-quality image generation of diffusion models, addressing the inherent challenges of facial attribute editing. By decoupling attribute editing from image generation, we overcome the inconsistency between the one-step adversarial learning in GANs and the multi-step denoising process in diffusion models. This decoupling, combined with feature-level adversarial training, enables flexible and controllable attribute manipulation without the computational burden of pixel-level adversarial training.
We further enhance style-attribute alignment through PriorMapper, which incorporates facial priors into style generation, and RefineExtractor, which utilizes Transformer-based global context modeling for improved style extraction. These advancements offer precise control over facial attributes while preserving non-target attributes.
Extensive experiments on CelebA-HQ demonstrate that AttDiff-GAN significantly outperforms existing methods in both visual quality and attribute editing accuracy, while effectively preserving irrelevant content. This framework offers a unique solution to facial attribute editing by combining the strengths of both GANs and diffusion models.
\bibliographystyle{IEEEtran}
\bibliography{reference}

\begin{thebibliography}{10}
\providecommand{\url}[1]{#1}
\csname url@samestyle\endcsname
\providecommand{\newblock}{\relax}
\providecommand{\bibinfo}[2]{#2}
\providecommand{\BIBentrySTDinterwordspacing}{\spaceskip=0pt\relax}
\providecommand{\BIBentryALTinterwordstretchfactor}{4}
\providecommand{\BIBentryALTinterwordspacing}{\spaceskip=\fontdimen2\font plus
\BIBentryALTinterwordstretchfactor\fontdimen3\font minus
  \fontdimen4\font\relax}
\providecommand{\BIBforeignlanguage}[2]{{%
\expandafter\ifx\csname l@#1\endcsname\relax
\typeout{** WARNING: IEEEtran.bst: No hyphenation pattern has been}%
\typeout{** loaded for the language `#1'. Using the pattern for}%
\typeout{** the default language instead.}%
\else
\language=\csname l@#1\endcsname
\fi
#2}}
\providecommand{\BIBdecl}{\relax}
\BIBdecl

\bibitem{WangCVPR2022}
T.~Wang, Y.~Zhang, Y.~Fan, J.~Wang, and Q.~Chen, ``High-fidelity gan inversion
  for image attribute editing,'' in \emph{IEEE/CVF Conference on Computer
  Vision and Pattern Recognition}, 2022, pp. 11\,379--11\,388.

\bibitem{PehlivanCVPR2023}
H.~Pehlivan, Y.~Dalva, and A.~Dundar, ``Styleres: Transforming the residuals
  for real image editing with stylegan,'' in \emph{IEEE/CVF Conference on
  Computer Vision and Pattern Recognition}, 2023, pp. 1828--1837.

\bibitem{HeTIP2019}
Z.~He, W.~Zuo, M.~Kan, S.~Shan, and X.~Chen, ``Attgan: Facial attribute editing
  by only changing what you want,'' \emph{IEEE Transactions on Image
  Processing}, vol.~28, no.~11, pp. 5464--5478, 2019.

\bibitem{HuangTCSVT2024}
W.~Huang, W.~Luo, X.~Cao, and J.~Huang, ``Interactive generative adversarial
  networks with high-frequency compensation for facial attribute editing,''
  \emph{IEEE Transactions on Circuits and Systems for Video Technology}, 2024.

\bibitem{DalvaTPAMI2023}
Y.~Dalva, H.~Pehlivan, O.~I. Hatipoglu, C.~Moran, and A.~Dundar,
  ``Image-to-image translation with disentangled latent vectors for face
  editing,'' \emph{IEEE Transactions on Pattern Analysis and Machine
  Intelligence}, 2023.

\bibitem{HuangAAAI2024}
W.~Huang, W.~Luo, J.~Huang, and X.~Cao, ``Sdgan: Disentangling semantic
  manipulation for facial attribute editing,'' in \emph{AAAI Conference on
  Artificial Intelligence}, vol.~38, no.~3, 2024, pp. 2374--2381.

\bibitem{ren2025facial}
F.~Ren, W.~Liu, F.~Wang, B.~Wang, and F.~Sun, ``Facial attribute editing via a
  balanced simple attention generative adversarial network,'' \emph{Expert
  Systems with Applications}, vol. 277, p. 127245, 2025.

\bibitem{RuizCVPR2024}
N.~Ruiz, Y.~Li, V.~Jampani, W.~Wei, T.~Hou, Y.~Pritch, N.~Wadhwa,
  M.~Rubinstein, and K.~Aberman, ``Hyperdreambooth: Hypernetworks for fast
  personalization of text-to-image models,'' in \emph{IEEE/CVF Conference on
  Computer Vision and Pattern Recognition}, 2024, pp. 6527--6536.

\bibitem{MengICLR2021}
C.~Meng, Y.~He, Y.~Song, J.~Song, J.~Wu, J.-Y. Zhu, and S.~Ermon, ``Sdedit:
  Guided image synthesis and editing with stochastic differential equations,''
  \emph{International Conference on Learning Representations}, 2021.

\bibitem{wang2025osdface}
J.~Wang, J.~Gong, L.~Zhang, Z.~Chen, X.~Liu, H.~Gu, Y.~Liu, Y.~Zhang, and
  X.~Yang, ``Osdface: One-step diffusion model for face restoration,'' in
  \emph{IEEE/CVF Conference on Computer Vision and Pattern Recognition}, 2025,
  pp. 12\,626--12\,636.

\bibitem{PreechakulCVPR2022}
K.~Preechakul, N.~Chatthee, S.~Wizadwongsa, and S.~Suwajanakorn, ``Diffusion
  autoencoders: Toward a meaningful and decodable representation,'' in
  \emph{IEEE/CVF Conference on Computer Vision and Pattern Recognition}, 2022,
  pp. 10\,619--10\,629.

\bibitem{KimCVPR2023}
G.~Kim, H.~Shim, H.~Kim, Y.~Choi, J.~Kim, and E.~Yang, ``Diffusion video
  autoencoders: Toward temporally consistent face video editing via
  disentangled video encoding,'' in \emph{IEEE/CVF Conference on Computer
  Vision and Pattern Recognition}, 2023, pp. 6091--6100.

\bibitem{KarrasICLR2018}
T.~Karras, ``Progressive growing of gans for improved quality, stability, and
  variation,'' \emph{International Conference on Learning Representations},
  2018.

\bibitem{KingmaarX2013}
D.~P. Kingma, ``Auto-encoding variational bayes,'' \emph{International
  Conference on Learning Representations}, 2014.

\bibitem{RezendearX2015}
D.~Rezende and S.~Mohamed, ``Variational inference with normalizing flows,'' in
  \emph{International Conference on Machine Learning}, 2015, pp. 1530--1538.

\bibitem{van2016pixel}
A.~Van Den~Oord, N.~Kalchbrenner, and K.~Kavukcuoglu, ``Pixel recurrent neural
  networks,'' in \emph{International Conference on Machine Learning}, 2016, pp.
  1747--1756.

\bibitem{GoodfellowNIPS2014}
I.~Goodfellow, J.~Pouget-Abadie, M.~Mirza, B.~Xu, D.~Warde-Farley, S.~Ozair,
  A.~Courville, and Y.~Bengio, ``Generative adversarial nets,'' \emph{Advances
  in Neural Information Processing Systems}, vol.~27, 2014.

\bibitem{KarrasCVPR2020}
T.~Karras, S.~Laine, M.~Aittala, J.~Hellsten, J.~Lehtinen, and T.~Aila,
  ``Analyzing and improving the image quality of stylegan,'' in \emph{IEEE/CVF
  Conference on Computer Vision and Pattern Recognition}, 2020, pp. 8110--8119.

\bibitem{10902474}
Z.~Luo, H.~Huang, L.~Yu, Y.~Li, B.~Zeng, and S.~Liu, ``Kernel reformulation
  with deep constrained least squares for blind image super-resolution,''
  \emph{IEEE Transactions on Circuits and Systems for Video Technology},
  vol.~35, no.~8, pp. 7380--7394, 2025.

\bibitem{10198457}
Y.~Lyu, Y.~Jiang, B.~Peng, and J.~Dong, ``Infostyler: Disentanglement
  information bottleneck for artistic style transfer,'' \emph{IEEE Transactions
  on Circuits and Systems for Video Technology}, vol.~34, no.~4, pp.
  2070--2082, 2024.

\bibitem{HoNIPS2020}
J.~Ho, A.~Jain, and P.~Abbeel, ``Denoising diffusion probabilistic models,''
  \emph{Advances in Neural Information Processing Systems}, vol.~33, pp.
  6840--6851, 2020.

\bibitem{SongICLR2021}
J.~Song, C.~Meng, and S.~Ermon, ``Denoising diffusion implicit models,''
  \emph{arXiv preprint arXiv:2010.02502}, 2020.

\bibitem{rombach2022high}
R.~Rombach, A.~Blattmann, D.~Lorenz, P.~Esser, and B.~Ommer, ``High-resolution
  image synthesis with latent diffusion models,'' in \emph{IEEE/CVF Conference
  on Computer Vision and Pattern Recognition}, 2022, pp. 10\,684--10\,695.

\bibitem{brooks2023instructpix2pix}
T.~Brooks, A.~Holynski, and A.~A. Efros, ``Instructpix2pix: Learning to follow
  image editing instructions,'' in \emph{IEEE/CVF Conference on Computer Vision
  and Pattern Recognition}, 2023, pp. 18\,392--18\,402.

\bibitem{11363596}
W.~Luo, S.~Yang, and H.~Niu, ``Soedit: Improving instruction-driven object
  editing by focusing on a single object within a cropped region,'' \emph{IEEE
  Transactions on Circuits and Systems for Video Technology}, pp. 1--1, 2026.

\bibitem{GaoCVPR2021}
Y.~Gao, F.~Wei, J.~Bao, S.~Gu, D.~Chen, F.~Wen, and Z.~Lian, ``High-fidelity
  and arbitrary face editing,'' in \emph{IEEE/CVF Conference on Computer Vision
  and Pattern Recognition}, 2021, pp. 16\,115--16\,124.

\bibitem{KarrasCVPR2019}
T.~Karras, ``A style-based generator architecture for generative adversarial
  networks,'' \emph{arXiv preprint arXiv:1812.04948}, 2019.

\bibitem{ShenTPAMI2020}
Y.~Shen, C.~Yang, X.~Tang, and B.~Zhou, ``Interfacegan: Interpreting the
  disentangled face representation learned by gans,'' \emph{IEEE Transactions
  on Pattern Analysis and Machine Intelligence}, vol.~44, no.~4, pp.
  2004--2018, 2020.

\bibitem{PatashnikCVPR2021}
O.~Patashnik, Z.~Wu, E.~Shechtman, D.~Cohen-Or, and D.~Lischinski, ``Styleclip:
  Text-driven manipulation of stylegan imagery,'' in \emph{IEEE/CVF
  International Conference on Computer Vision}, 2021, pp. 2085--2094.

\bibitem{WuCVPR2021}
Z.~Wu, D.~Lischinski, and E.~Shechtman, ``Stylespace analysis: Disentangled
  controls for stylegan image generation,'' in \emph{IEEE/CVF Conference on
  Computer Vision and Pattern Recognition}, 2021, pp. 12\,863--12\,872.

\bibitem{LiarX2016}
M.~Li, W.~Zuo, and D.~Zhang, ``Deep identity-aware transfer of facial
  attributes,'' \emph{arXiv preprint arXiv:1610.05586}, 2016.

\bibitem{ShenCVPR2017}
W.~Shen and R.~Liu, ``Learning residual images for face attribute
  manipulation,'' in \emph{IEEE/CVF Conference on Computer Vision and Pattern
  Recognition}, 2017, pp. 4030--4038.

\bibitem{dosovitskiy2020image}
A.~Dosovitskiy, L.~Beyer, A.~Kolesnikov, D.~Weissenborn, X.~Zhai,
  T.~Unterthiner, M.~Dehghani, M.~Minderer, G.~Heigold, S.~Gelly \emph{et~al.},
  ``An image is worth 16x16 words: Transformers for image recognition at
  scale,'' \emph{arXiv preprint arXiv:2010.11929}, 2020.

\bibitem{LiCVPR2021}
X.~Li, S.~Zhang, J.~Hu, L.~Cao, X.~Hong, X.~Mao, F.~Huang, Y.~Wu, and R.~Ji,
  ``Image-to-image translation via hierarchical style disentanglement,'' in
  \emph{IEEE/CVF Conference on Computer Vision and Pattern Recognition}, 2021,
  pp. 8639--8648.

\bibitem{shen2020interpreting}
Y.~Shen, J.~Gu, X.~Tang, and B.~Zhou, ``Interpreting the latent space of gans
  for semantic face editing,'' in \emph{IEEE/CVF Conference on Computer Vision
  and Pattern Recognition}, 2020, pp. 9243--9252.

\bibitem{HuangICCV2017}
X.~Huang and S.~Belongie, ``Arbitrary style transfer in real-time with adaptive
  instance normalization,'' in \emph{IEEE/CVF International Conference on
  Computer Vision}, 2017, pp. 1501--1510.

\bibitem{chen2025instruct}
S.~X. Chen, M.~Sra, and P.~Sen, ``Instruct-clip: Improving instruction-guided
  image editing with automated data refinement using contrastive learning,'' in
  \emph{IEEE/CVF Conference on Computer Vision and Pattern Recognition
  Conference}, 2025, pp. 28\,513--28\,522.

\bibitem{yang2021l2m}
G.~Yang, N.~Fei, M.~Ding, G.~Liu, Z.~Lu, and T.~Xiang, ``L2m-gan: Learning to
  manipulate latent space semantics for facial attribute editing,'' in
  \emph{IEEE/CVF Conference on Computer Vision and Pattern Recognition}, 2021,
  pp. 2951--2960.

\bibitem{HeuselNIPS2017}
M.~Heusel, H.~Ramsauer, T.~Unterthiner, B.~Nessler, and S.~Hochreiter, ``Gans
  trained by a two time-scale update rule converge to a local nash
  equilibrium,'' \emph{Advances in Neural Information Processing Systems},
  vol.~30, 2017.

\bibitem{HeCVPR2016}
K.~He, X.~Zhang, S.~Ren, and J.~Sun, ``Deep residual learning for image
  recognition,'' in \emph{IEEE/CVF Conference on Computer Vision and Pattern
  Recognition}, 2016, pp. 770--778.

\bibitem{saharia2022photorealistic}
C.~Saharia, W.~Chan, S.~Saxena, L.~Li, J.~Whang, E.~L. Denton, K.~Ghasemipour,
  R.~Gontijo~Lopes, B.~Karagol~Ayan, T.~Salimans \emph{et~al.},
  ``Photorealistic text-to-image diffusion models with deep language
  understanding,'' \emph{Advances in Neural Information Processing Systems},
  vol.~35, pp. 36\,479--36\,494, 2022.

\end{thebibliography}

\begin{IEEEbiography}
	[{\includegraphics[width=1in,height=1.25in,clip,keepaspectratio]{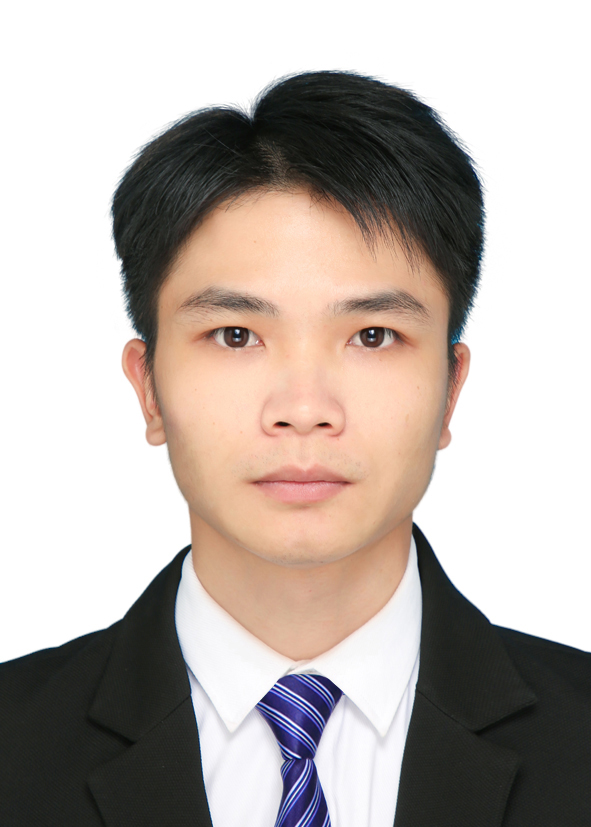}}]{Wenmin Huang}
	received the M.E. degree in intelligent science and technology from Tianjin Normal University, in 2021. He is currently working toward the Ph.D. degree at Sun Yat-sen University. His research interests include face attribute editing and computer vision.  
	He serves as a Program Committee for AAAI-2026 and a reviewer for IEEE Transactions on Circuits and Systems for Video Technology.
\end{IEEEbiography}

\begin{IEEEbiography}
	[{\includegraphics[width=1in,height=1.25in,clip,keepaspectratio]{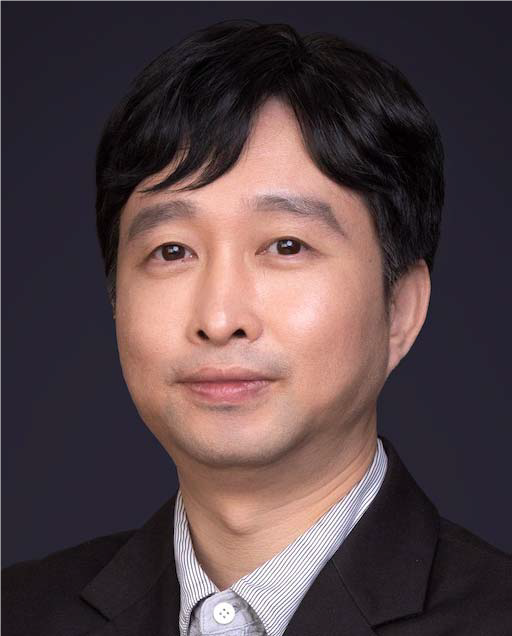}}]{Weiqi Luo} (Senior Member, IEEE) received his Ph.D. degree in 2008 from Sun Yat-sen University, Guangzhou, China. He is currently a Full Professor in the School of Computer Science and Engineering at Sun Yat-sen University. His research interests focus on digital multimedia forensics, steganography, and steganalysis. He serves as an Associate Editor for IEEE Transactions on Circuits and Systems for Video Technology, Information Forensics and Security, and Dependable and Secure Computing.
\end{IEEEbiography}	

\begin{IEEEbiography}
	[{\includegraphics[width=1in,height=1.25in,clip,keepaspectratio]{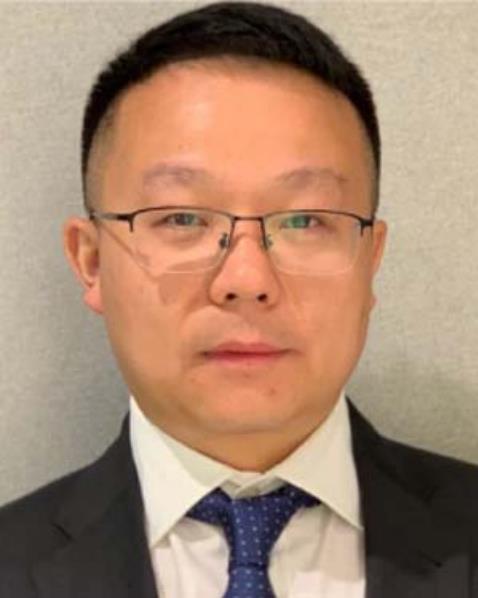}}]{Xiaochun Cao} (Senior Member, IEEE) received the B.E. and M.E. degrees in computer science from Beihang University (BUAA), China, and the Ph.D. degree in computer science from the University of Central Florida, USA. He is currently a Professor and the Dean of the School of Cyber Science and Technology, Shenzhen Campus of Sun Yat-sen University. His dissertation nominated for the university level Outstanding Dissertation Award. After graduation, he spent about three years with ObjectVideo Inc., as a Research Scientist. From 2008 to 2012, he was a Professor with Tianjin University. Before joining SYSU, he was a Professor with the Institute of Information Engineering, Chinese Academy of Sciences. He has authored or co-authored more than 200 journals and conference papers. In 2004 and 2010, he was a recipient of the Piero Zamperoni Best Student Paper Award at the International Conference on Pattern Recognition. He is on the editorial boards of IEEE Transactions on Pattern Analysis and Machine Intelligence and IEEE Transactions on Image Processing and was on the editorial boards of IEEE Transactions on Circuits and Systems for Video Technology and IEEE Transactions on Multimedia.
\end{IEEEbiography}	

\begin{IEEEbiography}
	[{\includegraphics[width=1in,height=1.25in,clip,keepaspectratio]{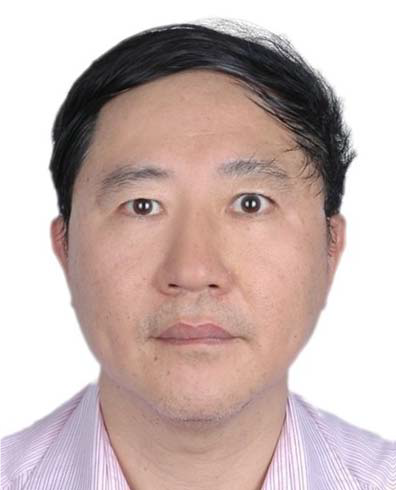}}]{Jiwu Huang}
	(Fellow, IEEE) received the B.S. degree from Xidian University, Xi’an, China, in 1982, the M.S. degree from Tsinghua University, Beijing, China, in 1987, and the Ph.D. degree from the Institute of Automation, Chinese Academy of Sciences, Beijing, in 1998. He is currently a Chief Professor with the Faculty of Engineering, Shenzhen MSU-BIT University, Shenzhen, 518116, China. He has coauthored more than 350 papers in important journals and conferences, with Google citations of more than 25000 and an H-index of 84. His current research interests include multimedia forensics and security.
\end{IEEEbiography}

\end{document}